\documentclass[3p,times,procedia]{elsarticle}
\usepackage{ecrc}
\usepackage[bookmarks=false]{hyperref}
    \hypersetup{colorlinks,
      linkcolor=blue,
      citecolor=blue,
      urlcolor=blue}

\volume{00}

\firstpage{1}

\journalname{Procedia Computer Science}

\runauth{author}

\jid{procs}

\usepackage{amssymb}

\usepackage[figuresright]{rotating}
\usepackage{times}
\usepackage{latexsym}
\usepackage{graphicx}
\usepackage{amsmath}
\usepackage{amssymb}
\usepackage{algorithm}
\usepackage{algorithmic}
\usepackage{latexsym}

\usepackage{fourier} 
\usepackage{array}
\usepackage{makecell}

\begin{document}
\begin{frontmatter}



\dochead{4th International Conference on Innovative Data Communication Technology and Application}

\title{Self-Enhancing Multi-filter Sequence-to-Sequence Model}


\author[a]{Yunhao Yang} 
\author[b]{Zhaokun Xue}
\author[a]{Andrew Whinston\corref{cor1}}

\address[a]{University of Texas at Austin, 2317 Speedway, Austin Texas 78712, USA}
\address[b]{University of Texas at Dallas, 800 W. Campbell Road, Richardson Texas 75080, USA}

\begin{abstract}
Representation learning is important for solving sequence-to-sequence problems in natural language processing. Representation learning transforms raw data into vector-form representations while preserving their features. However, data with significantly different features leads to heterogeneity in their representations, which may increase the difficulty of convergence.
We design a multi-filter encoder-decoder model to resolve the heterogeneity problem in sequence-to-sequence tasks.
The multi-filter model divides the latent space into subspaces using a clustering algorithm and trains a set of decoders (filters) in which each decoder only concentrates on the features from its corresponding subspace.
As for the main contribution, we design a self-enhancing mechanism that uses a reinforcement learning algorithm to optimize the clustering algorithm without additional training data. We run semantic parsing and machine translation experiments to indicate that the proposed model can outperform most benchmarks by at least 5\%.
We also empirically show the self-enhancing mechanism can improve performance by over 10\% and provide evidence to demonstrate the positive correlation between the model's performance and the latent space clustering.
\end{abstract}

\begin{keyword}
Recurrent neural network; LSTM; reinforcement learning; latent space clustering




\end{keyword}
\end{frontmatter}

\correspondingauthor[*]{Yunhao Yang. Tel.: +1-409-457-8367 ; fax: +1-512-741-0587.}
\email{yunhaoyang234@utexas.edu}


\vspace*{-6pt}

\section{Introduction}

A sequence-to-sequence (seq2seq) model takes in a sequence of tokens as input and construct an output sequence, where the length of the input sequence and the output sequence can be different. Machine translation and semantic parsing are two of the well-known seq2seq problems. Many works\cite{lstm, autoencoder_machine_translation, luong2015effective} use encoder-decoder framework to solve sequence-to-sequence problems. 

One of the challenges for the encoder-decoder models is that the training data with heterogeneous features may increase the difficulty of convergence. The model cannot concentrate on multiple heterogeneous features simultaneously: fitting one set of features may increase loss on another set of features.

To address this challenge, we apply a representation learning technique \cite{bengio2013representation} to divide heterogeneous features into clusters in the latent space, which is the final hidden space returned from the encoder \cite{yang2021training}. We first feed sequential data into the encoder to obtain the representations on the latent space. Second, we apply a clustering algorithm to divide the representations into clusters, each consisting of homogeneous features. Then, we construct multiple decoders to decode the representations in each cluster separately. Hence we can resolve the heterogeneity problem in seq2seq tasks. In addition to the multi-decoder architecture, we discover that the latent space clustering quality positively correlates to the model's performance. Therefore, enhancing the clustering algorithm can potentially improve the seq2seq model.

As for the main contribution to this work, we introduce a self-enhancing mechanism that uses a reinforcement learning \cite{rlbook} approach to enhance the clustering algorithm recursively. We apply a soft actor-critic algorithm to maximize the Silhouette coefficient of the clusters by optimizing the clustering algorithm. We expect the self-enhancing mechanism can continuously improve the model's performance without additional training data.

As a side contribution, we develop a neural network classifier to classify which cluster a given representation belongs to. Compared with traditional clustering algorithms such as K-means or the Gaussian mixture model, the neural network architecture can quickly adapt to our proposed self-enhancing mechanism. We gradually adjust the parameters in the neural network to obtain the optimal clusters.

In the experiments, we first show that the multi-filter architecture can indeed outperform the ordinary encoder-decoder model.
Then, we empirically show the positive correlation between the latent space clustering algorithm (evaluated by the Silhouette score) and the model's performance. Such observations indicate that the self-enhancing mechanism we proposed is effective.
\section{Related Work}

Encoder-decoder model is commonly used in solving sequence-to-sequence problems. Bidirectional LSTM \cite{lstm} with attention mechanism \cite{Dong_2016} is one of the state of the arts models. We build our LMS2S on top of this model and show how our model outperforms it.

Some existing works apply representation learning to the latent space. \cite{yang2021identifying} has shown that latent space representations can exploit the features and better preserve the critical attributes of the raw data.
\cite{bouchacourt2018multi} designs a multi-level latent space structure that can generate representation hierarchically. \cite{yang2017towards} divides the latent space into subspaces using a hard K-means clustering algorithm; \cite{jabi2019deep} introduces a soft K-means algorithm that enhances the results from \cite{yang2017towards}; \cite{dilokthanakul2016deep} develops a Gaussian mixture model to divide the latent space representations into a mixture of Gaussian distributions. The works \cite{Karrupusamy2020EffectiveTS, dhaya2021analysis} propose multi-filter frameworks for resolving heterogeneous image features.

All the works we mentioned above concentrate on processing images rather than language. They have shown that dividing the latent space into subspaces benefits performance. \cite{yang2021training} introduces for seq2seq tasks following these ideas. This work uses a Gaussian mixture model to divide the latent space into subspaces and apply multiple decoders to decode each subspace accordingly.
Our work starts from the architecture introduced in \cite{yang2021training}, modifies the clustering algorithm with a multi-layer perceptron, and presents a self-enhancing mechanism using reinforcement learning.

Reinforcement learning algorithms have been applied to enhance clustering results. Several works show that reinforcement learning can improve clustering algorithms \cite{Barbakh2007ClusteringWR, Bose2016SemiUnsupervisedCU} by designing proper reward functions. Unlike the existing works, we use a neural network classifier to assign clusters and apply a soft actor-critic algorithm to recursively optimize the parameters in the neural network.

\newcommand\numberthis{\addtocounter{equation}{1}\tag{\theequation}}

\section{Multi-filter Network Architecture}
The network consists of an encoder and a dummy decoder from an ordinary encoder-decoder model; a latent-space enhancer (multi-layer perceptron) that takes in the final hidden states from the encoder as inputs and projects them on the latent space; a cluster classifier for determining which cluster the latent space representation belongs to; certain number of decoders for various sets of features.

\begin{figure}[t]
    \begin{center}
        \includegraphics[width=0.7\linewidth]{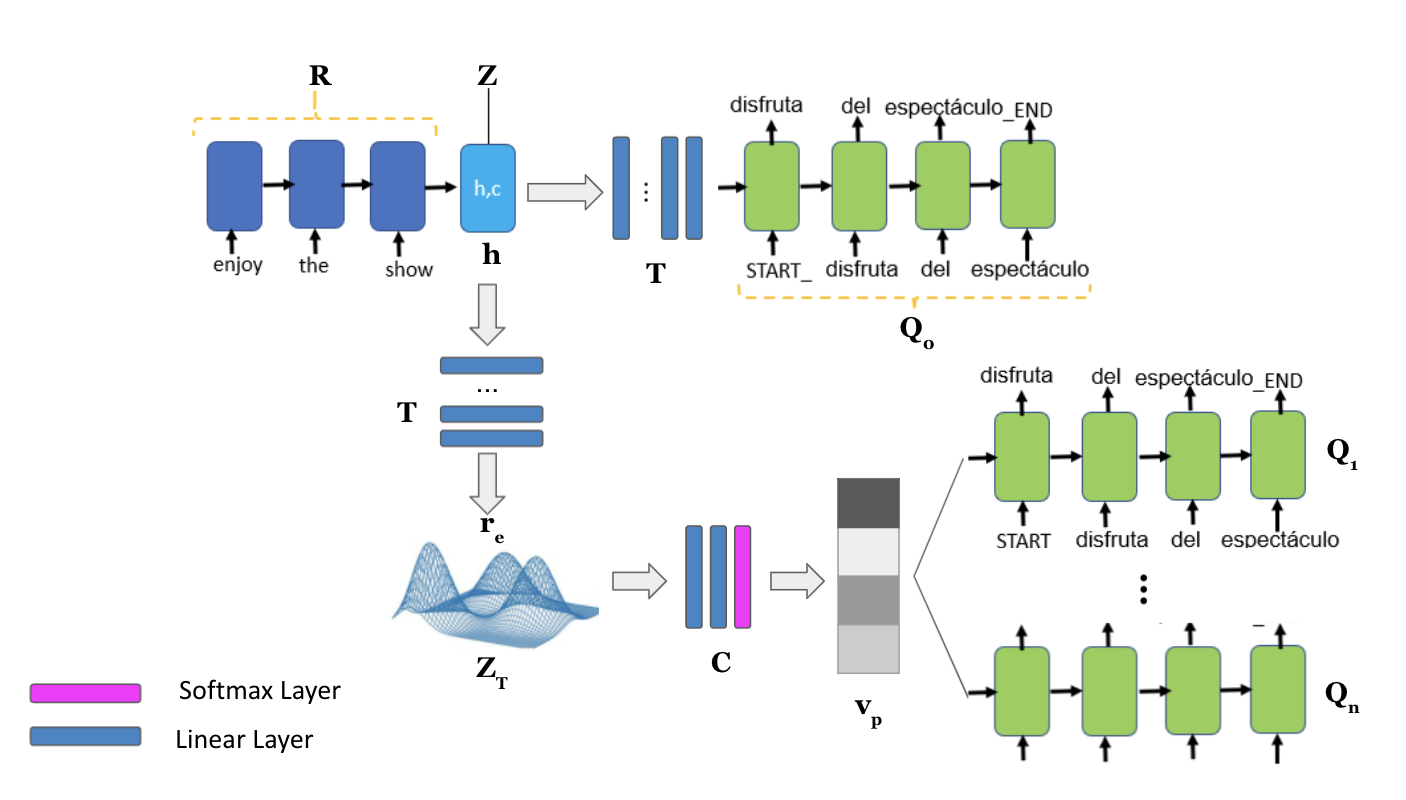}
    \end{center}
    \caption{Latent-enhanced Multi-filter Seq2seq Model Pipeline:
    we first use an encoder $R$ and a dummy decoder $Q_0$ to learn each representation $h$ at the latent space $Z$. Then, we apply an enhancer $T$ that transforms $h$ from $Z$ to $Z_T$. We denote the transformed $h$ on space $Z_T$ as the enhanced representation $r_e$. Following the enhancer, we apply a cluster classifier $C$ to assign each $r_e$ to a cluster. Each cluster $c_i$ corresponds to a filter $Q_i$ for decoding the representations from its correspondence cluster.
    }
    \label{fig:architecture}
\end{figure}

\subsection{Latent Enhanced Encoder-Decoder Model}
The latent enhanced encoder-decoder model consists of an encoder $R$ with parameters $\theta$, a dummy decoder $Q_0$ with parameters $\psi_0$, and an enhancer $T$ with parameters $\phi$ that connects to the encoder and projects the encoder outputs into the latent space. The encoder takes in a sequence of inputs $x$, then returns a sequence of outputs and a final hidden state $h$.
\begin{center}
    $h = h_{|x|}, \quad where \quad x_i, (h_i, c_i) = R( (x_{i-1}, h_{i-1}) | \theta), i \in \{1, 2, 3,...\}$,
\end{center}
where $x_i$, $h_i$, and $c_i$ are the $i^{th}$ output, hidden state, and cell state, respectively. $x_0$ is the pre-defined starting token, and $h_0$ and $c_0$ are randomly initialized. $|x|$ is the length of the output sequence $x$.

Each encoder and decoder consists of a bi-directional Long Short Term Memory (LSTM). LSTM falls into the category of recurrent neural networks (RNNs). Each LSTM cell has an input and output gate, similar to an RNN cell. However, traditional RNNs suffer from gradient exploding or vanishing during backpropagation. LSTM resolves this problem by adding a forget gate to control the gradient level. Every decoder consists of a dot-product attention mechanism \cite{luong2015effective}.

The encoder will generate a final hidden state after we feed in the entire input sequence. The final hidden states can be viewed as the representations of the input-output pairs. We denote the space where the final hidden states lie as $Z$. Then, we construct an enhancer and apply it in $Z$ to enhance the qualities of representations: 
$r_e = T(h|\phi)$.

The enhancer is a multi-layer perception that transforms the representation $h$ from $Z$ to a new space $Z_T$. We name the space $Z_T$ as the new latent space. The enhanced representation, denoted as $r_e$, is fed as the input of the first decoder cell:
\begin{center}
    $y_1 = Q_0( (x_0, r_e) | \psi), \quad y_i = Q_0( (y_{i-1}, r_{i-1}) | \psi), i = \{1,2,3,...\}$.    
\end{center}

In addition, the $r_e$ is also fed as the input of the cluster classifier $C$ for cluster assignment, which we will discuss in the later section.

We use a Negative Likelihood Loss to optimize the trainable weights $\theta$, $\phi$, and $\psi$. Let $\overline{y}$ be the output vector sequence of the network and $y$ be the label sequence. $\overline{y}$ is a sequence of $N$-dimensional probability vectors and $N$ is the dictionary size. $y$ is a sequence of $N$ positive integers. We denote the $j^{th}$ index of the $i^{th}$ element in $\overline{y}$ as $\overline{y}_{(i, j)}$.
Then, we can compute the loss as the following:
\begin{equation}
    \label{eq:nll}
    \mathcal{L}(\overline{y}, y) = \sum_{i=1}^N l_i, \quad l_i = -\overline{y}_{(i, y_i)}.
\end{equation}

After the model is converged, we remove the dummy decoder and only keep the encoder and the enhancer. Then, we fix the parameters in the encoder and the enhancer in the following steps.

\subsection{Cluster Assignment}
We construct a cluster classifier $C$ with parameters $\omega$ to assign the enhanced representations into clusters. The classifier $C$ takes $r_e$ as input and output a probability vector $v_c$ for cluster assignment: $v_p = C(r_e | \omega)$.
The classifier $C$ consists of two linear layers and a softmax layer. The output of $C$ captures the probability of a given data belonging to a cluster:
\begin{center}
    $v_p[i] = \mathbb{P}[T(h_x|\phi) \in c_i]$.
\end{center}
We can take the cluster $c_j$ with the highest probability as the cluster the sample representation $x$ belongs to:
\begin{center}
    $T(h_x|\phi) \in c_{argmax(v_p)}$.
\end{center}
Then, the enhanced representation $T(h_x|\phi)$ can be fed into the filter $Q_j$ that corresponding to $c_j$. Hence the output can be constructed.

The parameters in $C$ will not be adjusted by any loss computed from the decoder's outputs. Instead, we applied a reinforcement learning algorithm to optimize the parameters in $C$, which is discussed in the later section. 


\subsection{Training Decoders}
We construct a set of decoders $Q = \{Q_i, i = 1,...,n\}$ to construct output sequences from the latent space representations. We apply the clustering algorithm to divide the enhanced latent space $Z_T$ into subspaces. Then, we assign a decoder from the decoder set $Q$ to each subspace. Each decoder concentrates on the data within the subspace it is assigned to. All the decoders are identical to the dummy decoder. However, to distinguish from the dummy decoder $Q_0$, we denote the decoders from set $Q$ as \textit{filters}.

In the training stage, we apply the cluster classifier $C$ to divide the data into $n$ clusters. We set up $n$ filters corresponding to $n$ clusters. All the filters are initially the same, but we optimize their trainable weights independently. We extract the data records from cluster $c_i$ to train the filter $Q_i$.
We use the Negative Likelihood Loss as presented in Equation \ref{eq:nll} to update the weights in the filters. Note that the NLLoss for each filter is computed separately; hence the gradient updates for each filter are independent.
Through the entire training procedure, the parameters in the encoder $R$ and the enhancer $T$ are fixed; thus they will not be updated. 

In the evaluation stage, we first encode and obtain the representation $r_e$ from the latent space $Z_T$. Then, we apply the cluster classifier $C$ to determine which cluster the input data belongs to. Once the cluster $c_i$ is determined, we use the corresponding filter $Q_i$ to construct the output sequence.

\section{Self-Enhancing Algorithm}
\label{sec:lea}

\begin{figure}[t]
    \centering
    \includegraphics[width=0.6\linewidth]{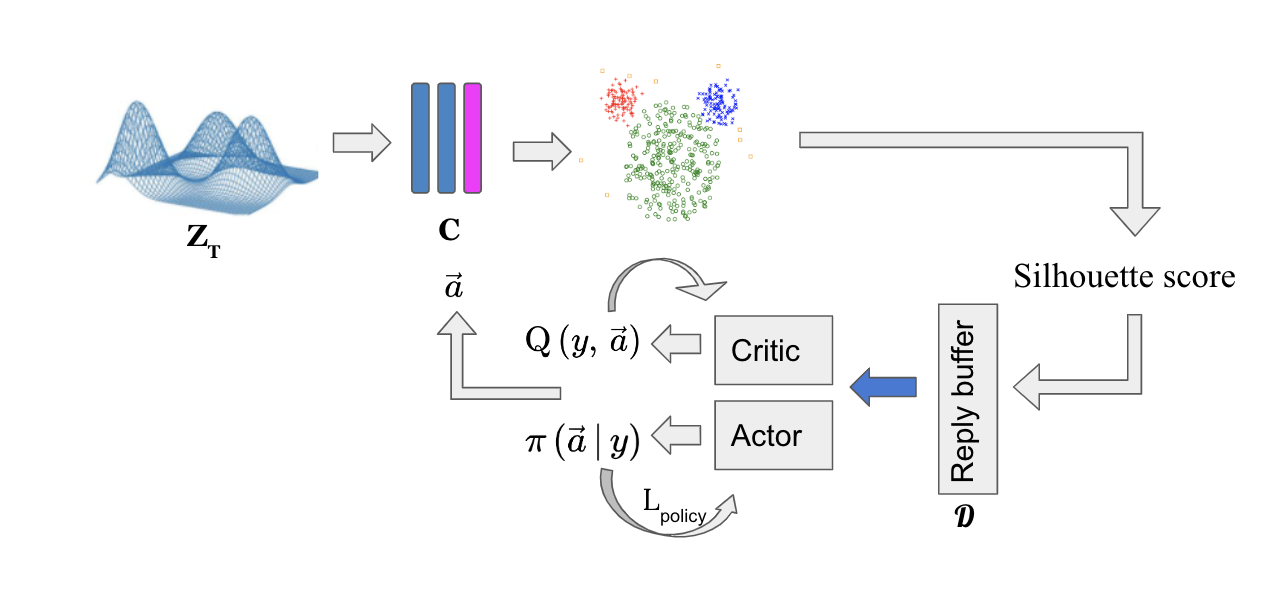}
    \caption{Self-enhancing Pipeline}
    \label{fig:self}
\end{figure}

To optimize the latent space, we introduce a Soft Actor-Critic (SAC)\cite{haarnoja2018soft} reinforcement learning algorithm that is utilized for optimizing the trainable parameters $\omega$ in the cluster classifier $C$. To adapt the SAC algorithm, we need to define the state space, action space, reward function, and terminating state.

The state is the parameters $\omega$, which is a set of matrices. The actions space is a set of matrices whose dimensions are identical to the state matrices. We consider the action as a set of weights multiplied to the state matrices:
\begin{equation}
    \omega_i' = a_i \cdot \omega_i
    \label{eq:action}
\end{equation}
where $a$ is an action, $a_i$ is a single weight matrix multiplied to the $i^{th}$ parameter $\omega_i$.

We compute the Silhouette Coefficient $S_c$ of the clusters assigned by $C$ and use it for the reward:
\begin{equation}
    R = k \cdot S_c + b
    \label{eq:reward}
\end{equation}
where $k$ and $b$ are the user-defined constants.

The reinforcement learning is applied after the encoder, and the dummy decoder is fine-tuned prior to training the filters. We set a target Silhouette Coefficient $S_c^{target}$, and a maximum number of steps allowed $S_{max}$. We define the terminating state $\mathcal{T}$ as the following:
\begin{center}
    $\mathcal{T} = t < S_{max} \& S_t < S_c^{target}$,
\end{center}
where $t$ is the current timestamp and $S_t$ is the Silhouette Coefficient at timestamp $t$.
We expect the SAC algorithm can enhance the clustering algorithm by helping it achieve a high Silhouette Coefficient. After the SAC algorithm improves the clustering quality, we can start training the multiple filters.

\section{Experiments}
In this section, we will first prove the effectiveness of the multi-filter architecture. To achieve this, we conduct several comparative experiments. These experiments compare the performance of our latent-enhanced multi-filter model with the traditional encoder-decoder model, as well as some other baselines. The experiments are conducted on two of the classical sequence-to-sequence tasks- semantic parsing and machine translation.
Second, we will demonstrate the positive correlation between the model's performance and the quality of the latent space clustering and indicate the necessity of the self-enhancing algorithm.

\subsection{Performance Examination on Semantic Parsing}

\begin{figure}[t]
    \centering
    \includegraphics[width=0.24\linewidth]{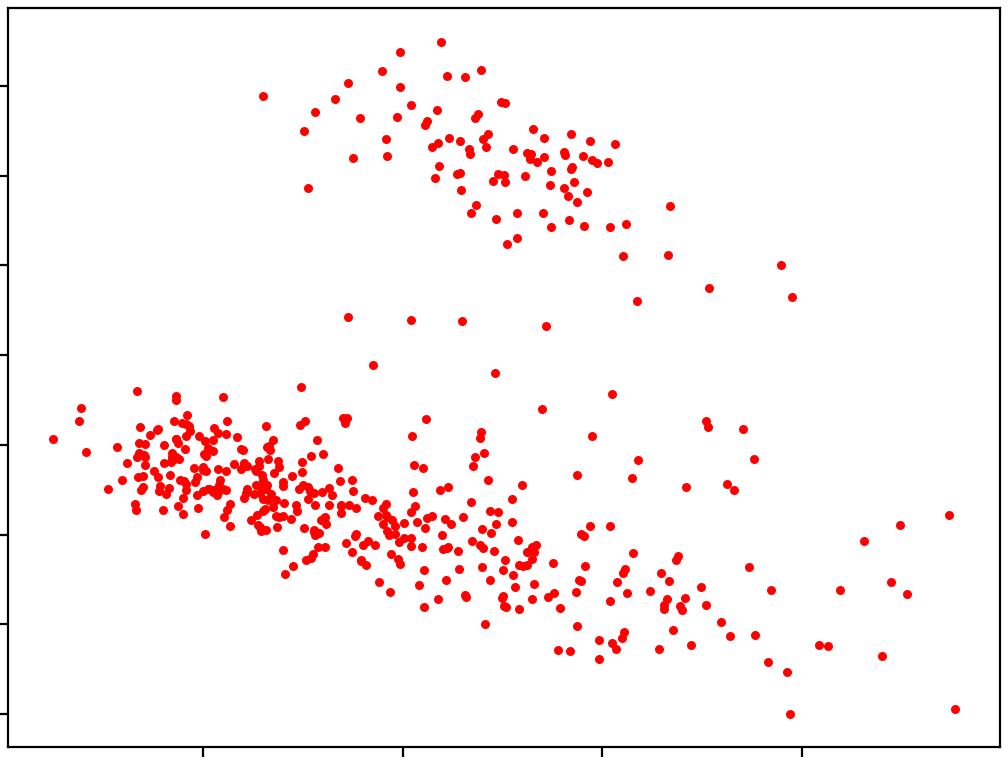}
    \includegraphics[width=0.24\linewidth]{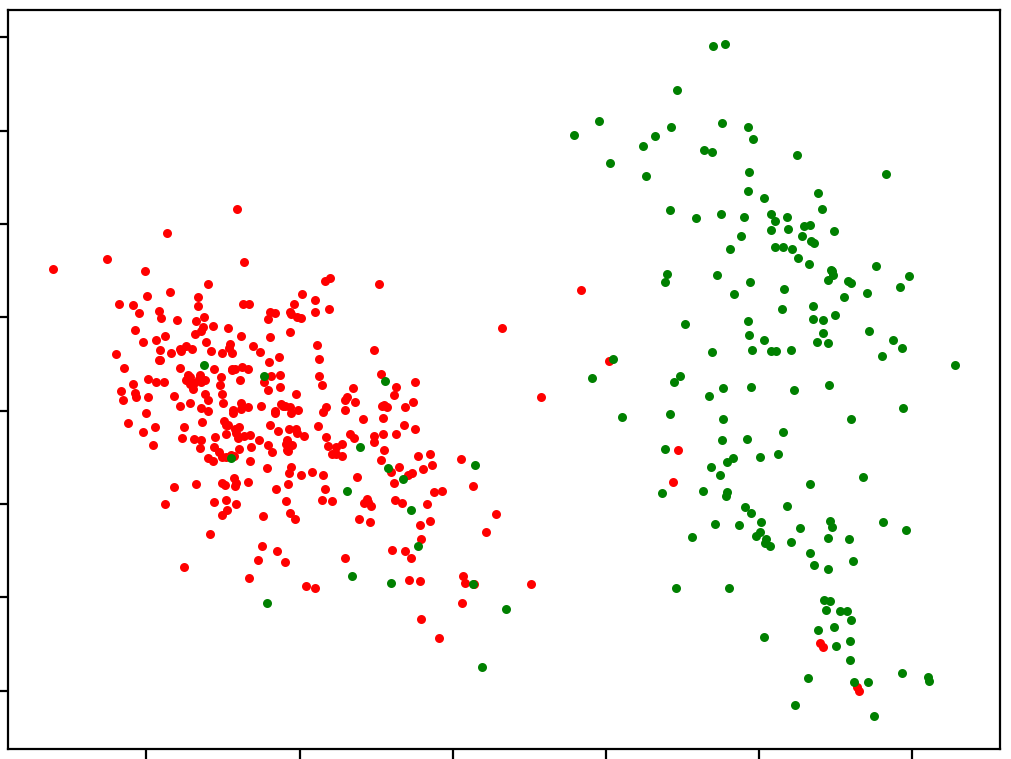}
    \includegraphics[width=0.24\linewidth]{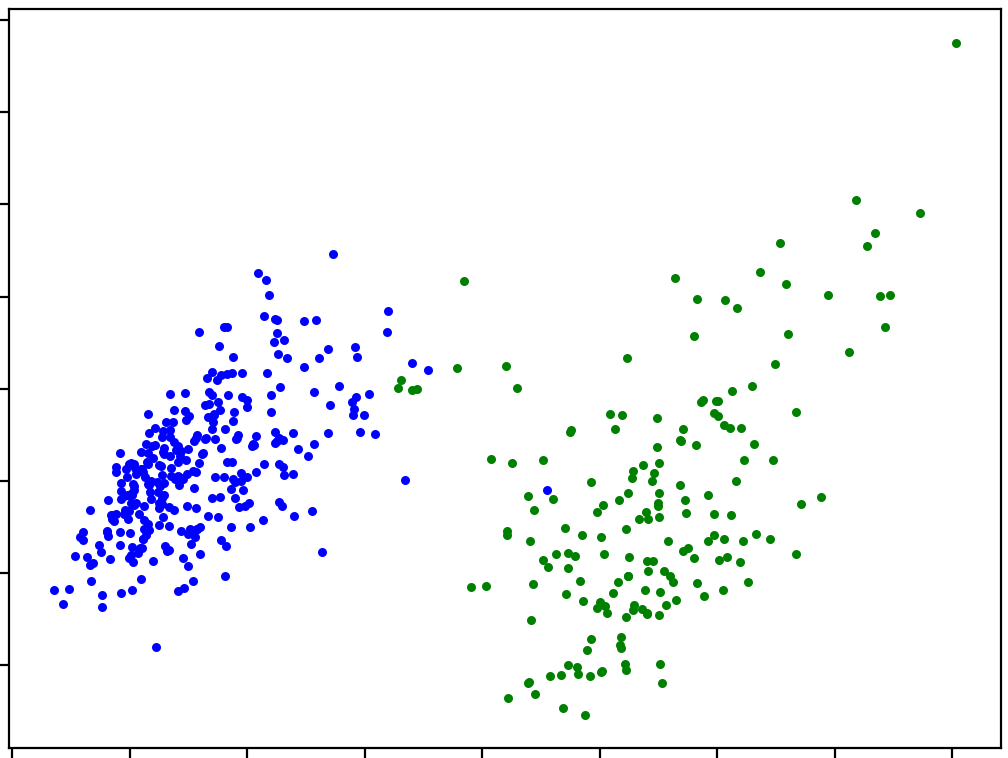}
    \includegraphics[width=0.24\linewidth]{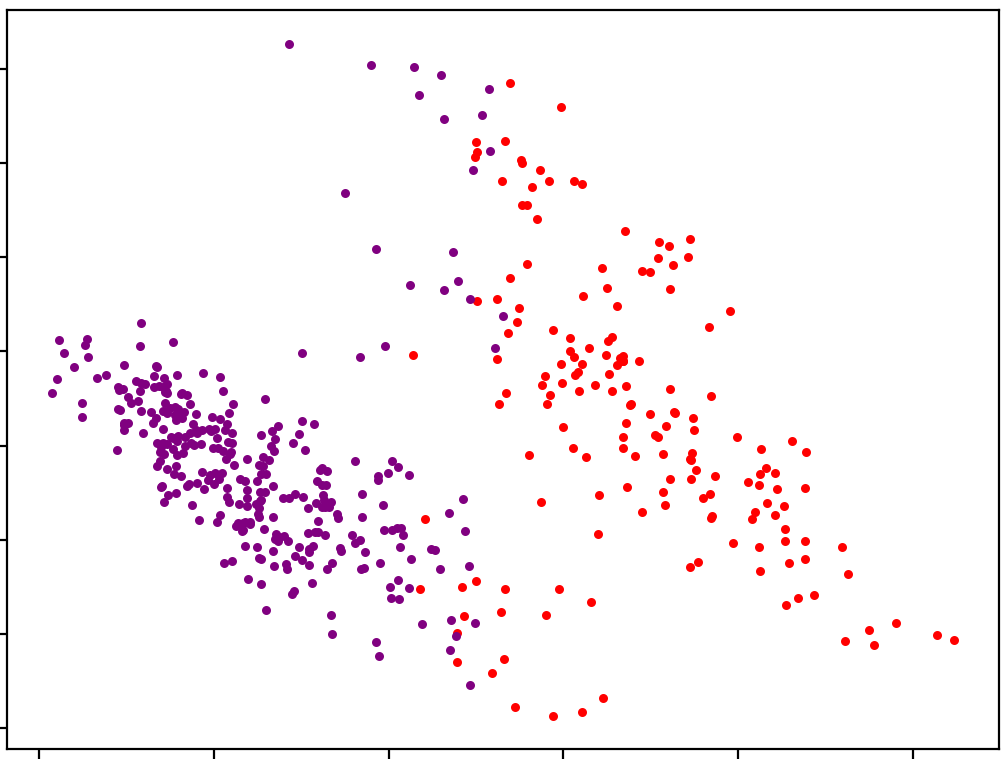}
    \caption{latent space clustering in Geo-query training data. Plots from left to right show the clustering results for two, three, and four clusters, respectively.}
    \label{fig:geo}
\end{figure}

We evaluate the proposed architecture on a semantic parsing task, where we use the Geo-query dataset \cite{zelle_mooney} that contains a set of geographical questions. We use the token level accuracy, and the denotation accuracy \cite{Jia_2016, liang-etal-2011-learning} to quantify the performance of this architecture.

We implemented the proposed architecture in PyTorch with specifications stated in \ref{tab:specification}.

To enhance the latent space clustering, we utilize multi-layer perceptron (MLP) as the policy for SAC. In the SAC algorithm, we set the max number of learning step to 500, $k=100$ and $b=25$ in Equation \ref{eq:reward}, target Silhouette score $S_c^{target} = 0.55$, and keep the default settings from the SAC implementation from Stable-Baselines3\cite{stable-baselines3}.

\begin{table}[!ht]
    \centering
    \caption{Experimental Specifications: we observed that our model can converge within 10 epochs under the current hyper-parameter settings.}
    \begin{tabular}{||c c||}
     \hline
         Training size & Validation size \\
         480 & 120 \\
         \hline
          Optimizer & Learning rate \\
          Adam & 0.001 \\
          \hline
         Epoch number & Dropout rate \\
          10 & 0.2 \\
          \hline
         Hidden dimension & Latent dimension\\ 
         200 & 200 \\
         \hline
         Embedding dimension & Bidirectional \\
           150 & True\\
          \hline
    \end{tabular}
    \label{tab:specification}
\end{table}


Figure \ref{fig:geo} shows the latent space clustering results on different number of clusters after the classifier is optimized by the SAC algorithm. The interesting observation is that every cluster classifier generates two clusters, regardless of the filter number we assign to the model. This observation indicates that the optimal number of clusters in the latent space is two in this task. And the SAC reinforcement learning algorithm optimizes the weights in the cluster classifier. Hence if we set the number of clusters to greater than two, the cluster classifier will generate empty clusters and ensure there are only two clusters that are non-empty. Because the SAC algorithm has learned that this way of clustering (separating into two clusters) will maximize the Silhouette score. Our 2-filters LES2S achieves the best performance over both metrics as shown in Table \ref{tab:1}.

\begin{table}[!htbp]
\caption{Performance comparison between the baselines \cite{Dong_2016} and LMS2S model, in terms of Token-level accuracy and Denotation accuracy. The ordinary encoder-decoder model is denoted as Enc-Dec. Note that some of the baselines are not reporting denotation accuracy.}
\begin{center}
 \begin{tabular}{||c c c||} 
 \hline
 Model & Token-Level & Denotation \\
 \hline
 Enc-Dec & 77.4 & 43.8 \\
 \thead{SCISSOR  \cite{10.5555/1706543.1706546}} & 72.3 &  \\
 \thead{KRISP  \cite{kate-mooney-2006-using}} & 71.7 &  \\
 \thead{WASP \cite{wong-mooney-2006-learning}} & 74.8 &  \\
 \thead{LEMS \cite{yang2021training}} & 78.3 & 50.9  \\
 \thead{ZC05  \cite{zelle_mooney}} & 79.3 & \\
 \textbf{LMS2S} & \textbf{81.7} & \textbf{60.8} \\
 \hline
\end{tabular}
\end{center}
\label{tab:1}
\end{table}


\subsection{Performance Examination on Machine Translation}
We also justify the performance of the proposed architecture in machine translation tasks using the Multi30k English-French dataset\cite{multi30k}. This dataset is split to 29,000 training data records and 1,000 testing data records. The experimental setups are identical to semantic parsing. We use BLEU score \cite{papineni-etal-2002-bleu} with  $N = 4$ and uniform weights $w_n = 1/4$ to evaluate the translation results. The BLEU score range is from 0 to 100, where 100 indicates a perfect match.



\begin{figure}[t]
    \centering
    \includegraphics[width=0.24\linewidth]{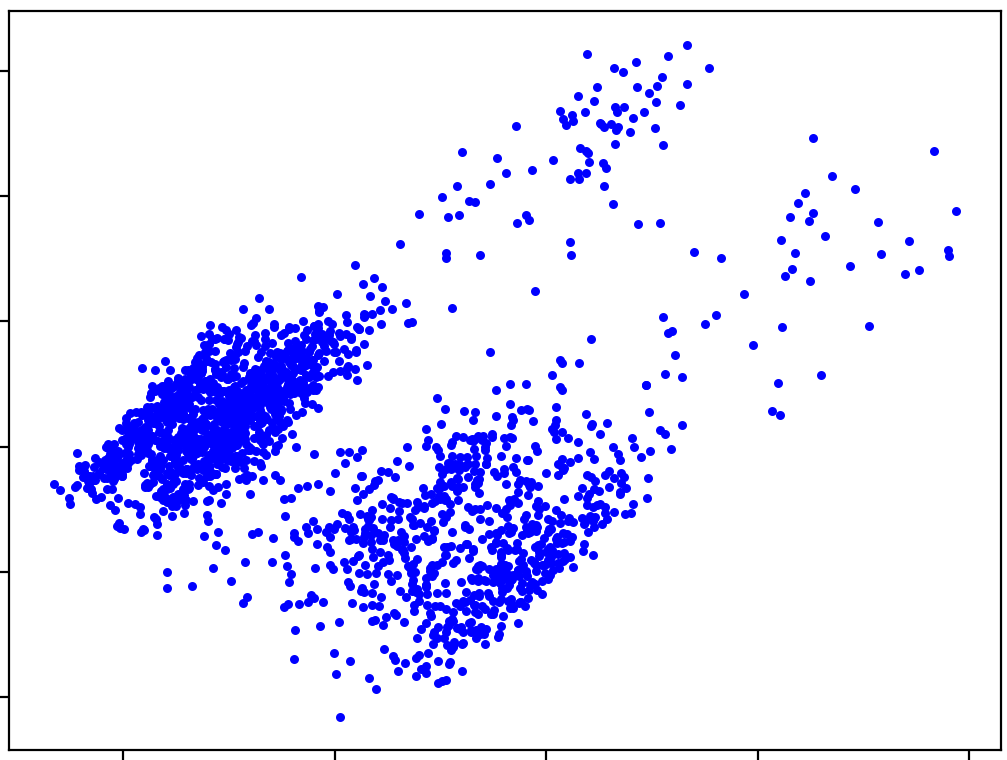}
    \includegraphics[width=0.24\linewidth]{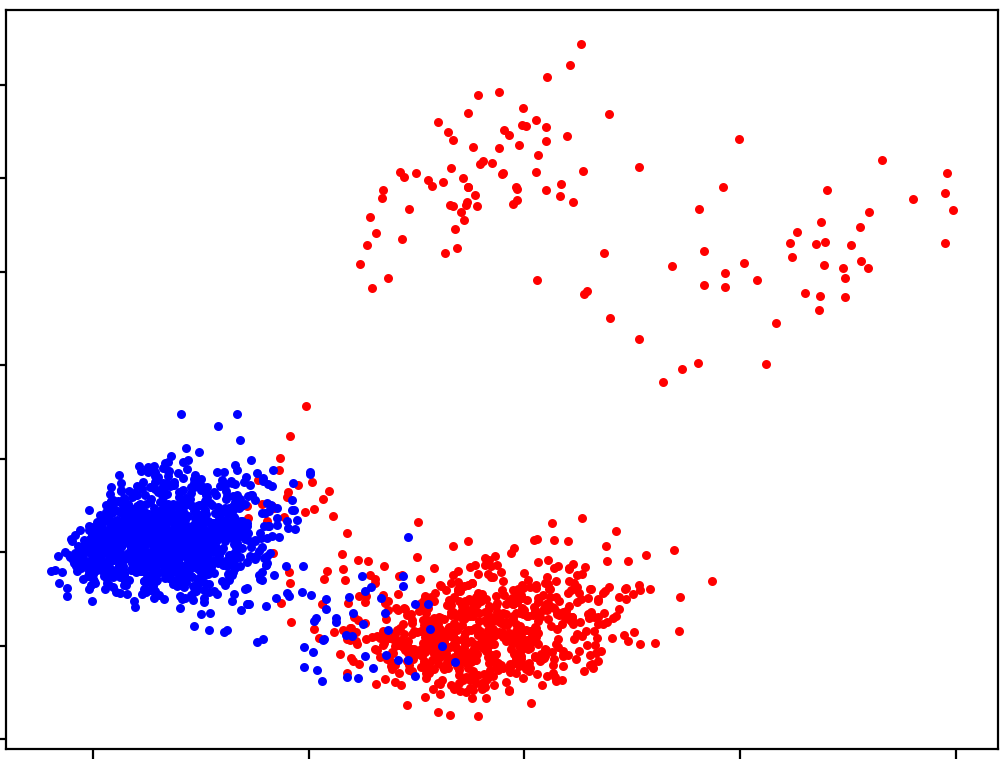}
    \includegraphics[width=0.24\linewidth]{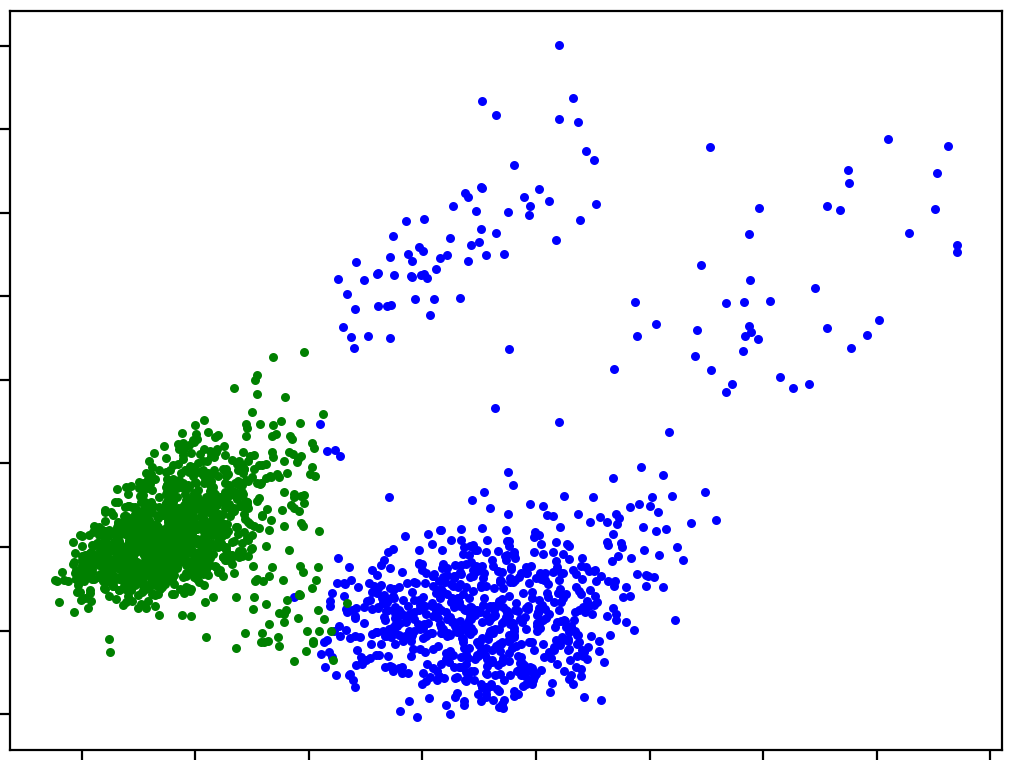}
    \includegraphics[width=0.24\linewidth]{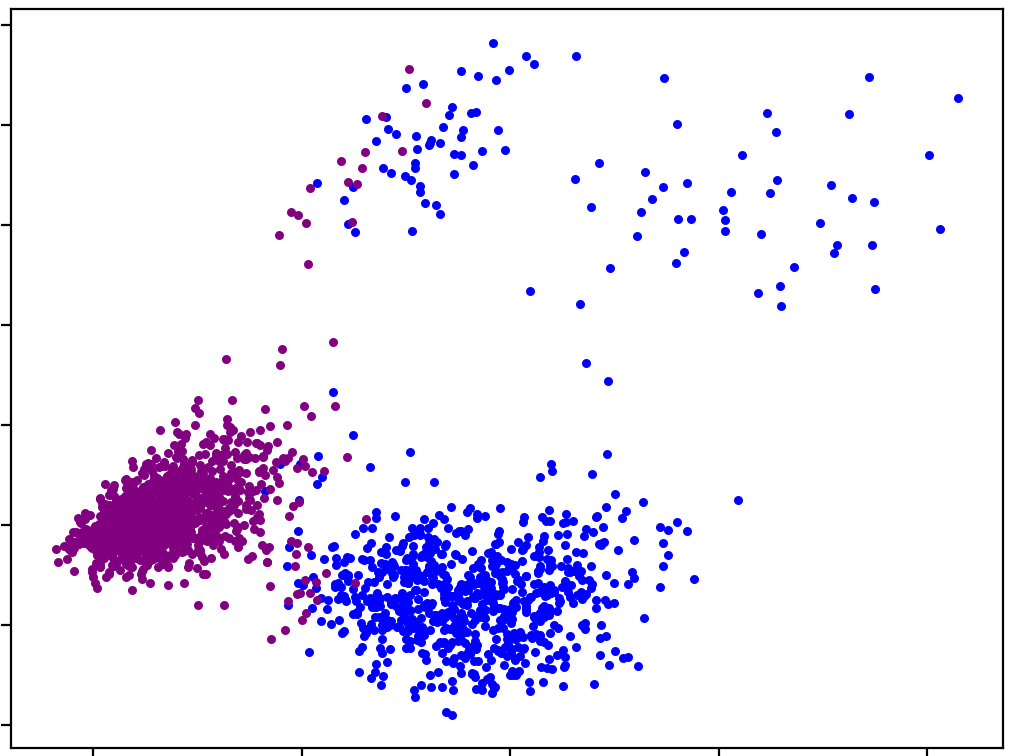}
    \caption{The figures show the clustering results on the latent space representations of the Multi-30K dataset. Figures left to right show the results on 1 to 4 clusters, respectively.}
    \label{fig:trans}
\end{figure}

As shown in Figure \ref{fig:trans}, the RL agent learns that the Silhouette score can be maximized when the latent space representations are separated into two clusters. Therefore, even if we increase the number of clusters, the classifier will form empty clusters so that there are still two non-empty clusters. Then, we can use a 2-filter LES2S model for the following experiments.

\begin{table}[!htbp]
\caption{Performance comparison between our LMS2S model and several baselines in the machine translation task.}
\begin{center}
 \begin{tabular}{||c c||} 
 \hline
 Model & BLEU\\
 \hline
 Baseline (text-only NMT) & 44.3\\
 \thead{SHEF \_ShefClassProj\_C \cite{Elliott_2017}} & 43.6 \\
 \thead{LEMS \cite{yang2021training}} & 46.3  \\
 \thead{CUNI Neural Monkey Multimodel MT\_C \cite{NeuralMonkey:2017}}
  & 49.9 \\
 
 \thead{LIUMCVC\_NMT\_C \cite{Caglayan_2017}}  & 53.3 \\
 
 \thead{DCU-ADAPT MultiMT C \cite{Elliott_2017}}  & 54.1 \\
 
 \textbf{LES2S} & \textbf{55.7} \\
 \hline
\end{tabular}
\end{center}
\label{tab:2}
\end{table}

Table \ref{tab:2} lists the BLUE scores of the baseline models and our LES2S model. Among these results, our model achieves the best performance. Moreover, the comparative experiment between the ordinary encoder-decoder model and our model also shows the effectiveness of the multi-filter architecture.

\subsection{Self-Enhancing Algorithm}

In both tasks, we use the soft actor-critic (SAC) algorithm to enhance the clustering quality. The results have shown that the SAC algorithm is able to improve the Silhouette score (generate better clusters).

To explore how the Silhouette score affects the model's performance, we train a set of LES2S models for each task. Every model has been trained under the same hyper-parameter settings as stated in Table \ref{tab:specification}.

By setting the learning steps in the SAC algorithm to 10, 20, 30, 50, 100, 200, 300, and 500, respectively, we can get clustering results in terms of Silhouette scores. Then, we compare the models' performances with Silhouette scores to show the significance of the latent space clustering.



\begin{figure}[t]
    \centering
    \includegraphics[width=0.5\linewidth]{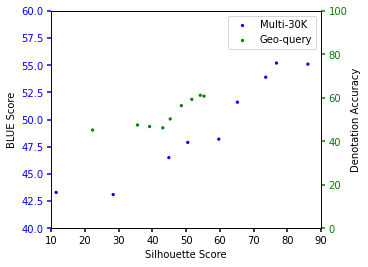}
    \caption{The figure shows how the model's performance improves as the clustering algorithm is enhancing.}
    \label{fig:rl}
\end{figure}

Figure \ref{fig:rl} shows the model's performance vs. Silhouette scores. We can observe that the Silhouette score is positively correlated to both of the evaluation metrics of the two tasks. These results show that we can improve the model's performance by optimizing the latent space clustering. Hence we have proved the significance of the latent space clustering.

\section{Conclusion}

In this work, we show that the multi-filter Seq2Seq model can resolve the problem of heterogeneous features in the dataset. We provide empirical evidence to demonstrate the positive correlation between the clustering quality and the model's performance. Therefore, we introduce a self-enhancing mechanism that uses a soft actor-critic reinforcement learning algorithm to optimize the clustering quality. The mechanism can improve the language model's performance without adding more training data. The experiments show that the self-enhancing mechanism can improve the clustering quality and the model's performance.

\paragraph{Limitation and Future Direction}
Same with the majority of the reinforcement learning algorithms, the proposed self-enhancing mechanism has difficulties with its convergence, especially when the training dataset gets more extensive. As a future direction, we will explore more up-to-date reinforcement learning algorithms such as Proximal Policy Optimization (PPO) and construct reward functions more related to the actions.


\bibliography{ref}

\begin{thebibliography}{29}
\expandafter\ifx\csname natexlab\endcsname\relax\def\natexlab#1{#1}\fi
\providecommand{\url}[1]{\texttt{#1}}
\providecommand{\href}[2]{#2}
\providecommand{\path}[1]{#1}
\providecommand{\DOIprefix}{doi:}
\providecommand{\ArXivprefix}{arXiv:}
\providecommand{\URLprefix}{URL: }
\providecommand{\Pubmedprefix}{pmid:}
\providecommand{\doi}[1]{\href{http://dx.doi.org/#1}{\path{#1}}}
\providecommand{\Pubmed}[1]{\href{pmid:#1}{\path{#1}}}
\providecommand{\bibinfo}[2]{#2}
\ifx\xfnm\relax \def\xfnm[#1]{\unskip,\space#1}\fi
\bibitem[{Barbakh and Fyfe(2007)}]{Barbakh2007ClusteringWR}
\bibinfo{author}{Barbakh, W.}, \bibinfo{author}{Fyfe, C.},
  \bibinfo{year}{2007}.
\newblock \bibinfo{title}{Clustering with reinforcement learning}, in:
  \bibinfo{booktitle}{International Conference on Intelligent Data Engineering
  and Automated Learning}, \bibinfo{organization}{Springer}. pp.
  \bibinfo{pages}{507--516}.
\bibitem[{Bengio et~al.(2013)Bengio, Courville and
  Vincent}]{bengio2013representation}
\bibinfo{author}{Bengio, Y.}, \bibinfo{author}{Courville, A.},
  \bibinfo{author}{Vincent, P.}, \bibinfo{year}{2013}.
\newblock \bibinfo{title}{Representation learning: A review and new
  perspectives}.
\newblock \bibinfo{journal}{IEEE transactions on pattern analysis and machine
  intelligence} \bibinfo{volume}{35}, \bibinfo{pages}{1798--1828}.
\bibitem[{Bose and Huber(2016)}]{Bose2016SemiUnsupervisedCU}
\bibinfo{author}{Bose, S.}, \bibinfo{author}{Huber, M.}, \bibinfo{year}{2016}.
\newblock \bibinfo{title}{Semi-unsupervised clustering using reinforcement
  learning}.
\bibitem[{Bouchacourt et~al.(2018)Bouchacourt, Tomioka and
  Nowozin}]{bouchacourt2018multi}
\bibinfo{author}{Bouchacourt, D.}, \bibinfo{author}{Tomioka, R.},
  \bibinfo{author}{Nowozin, S.}, \bibinfo{year}{2018}.
\newblock \bibinfo{title}{Multi-level variational autoencoder: Learning
  disentangled representations from grouped observations}, in:
  \bibinfo{editor}{McIlraith, S.A.}, \bibinfo{editor}{Weinberger, K.Q.} (Eds.),
  \bibinfo{booktitle}{Proceedings of the Thirty-Second {AAAI} Conference on
  Artificial Intelligence, (AAAI-18), the 30th innovative Applications of
  Artificial Intelligence (IAAI-18), and the 8th {AAAI} Symposium on
  Educational Advances in Artificial Intelligence (EAAI-18), New Orleans,
  Louisiana, USA, February 2-7, 2018}, \bibinfo{publisher}{{AAAI} Press}. pp.
  \bibinfo{pages}{2095--2102}.
\bibitem[{Caglayan et~al.(2017)Caglayan, Aransa, Bardet, García-Martínez,
  Bougares, Barrault, Masana, Herranz and van~de Weijer}]{Caglayan_2017}
\bibinfo{author}{Caglayan, O.}, \bibinfo{author}{Aransa, W.},
  \bibinfo{author}{Bardet, A.}, \bibinfo{author}{García-Martínez, M.},
  \bibinfo{author}{Bougares, F.}, \bibinfo{author}{Barrault, L.},
  \bibinfo{author}{Masana, M.}, \bibinfo{author}{Herranz, L.},
  \bibinfo{author}{van~de Weijer, J.}, \bibinfo{year}{2017}.
\newblock \bibinfo{title}{Lium-cvc submissions for wmt17 multimodal translation
  task}.
\newblock \bibinfo{journal}{Proceedings of the Second Conference on Machine
  Translation} \URLprefix \url{http://dx.doi.org/10.18653/v1/W17-4746},
  \DOIprefix\doi{10.18653/v1/w17-4746}.
\bibitem[{Cvetko(2020)}]{autoencoder_machine_translation}
\bibinfo{author}{Cvetko, T.}, \bibinfo{year}{2020}.
\newblock \bibinfo{title}{Autoencoders for translation}.
\bibitem[{Dhaya(2021)}]{dhaya2021analysis}
\bibinfo{author}{Dhaya, R.}, \bibinfo{year}{2021}.
\newblock \bibinfo{title}{Analysis of adaptive image retrieval by transition
  kalman filter approach based on intensity parameter}.
\newblock \bibinfo{journal}{Journal of Innovative Image Processing (JIIP)}
  \bibinfo{volume}{3}, \bibinfo{pages}{7--20}.
\bibitem[{Dilokthanakul et~al.(2016)Dilokthanakul, Mediano, Garnelo, Lee,
  Salimbeni, Arulkumaran and Shanahan}]{dilokthanakul2016deep}
\bibinfo{author}{Dilokthanakul, N.}, \bibinfo{author}{Mediano, P.A.},
  \bibinfo{author}{Garnelo, M.}, \bibinfo{author}{Lee, M.C.},
  \bibinfo{author}{Salimbeni, H.}, \bibinfo{author}{Arulkumaran, K.},
  \bibinfo{author}{Shanahan, M.}, \bibinfo{year}{2016}.
\newblock \bibinfo{title}{Deep unsupervised clustering with gaussian mixture
  variational autoencoders}.
\newblock \bibinfo{journal}{arXiv preprint arXiv:1611.02648} .
\bibitem[{Dong and Lapata(2016)}]{Dong_2016}
\bibinfo{author}{Dong, L.}, \bibinfo{author}{Lapata, M.}, \bibinfo{year}{2016}.
\newblock \bibinfo{title}{Language to logical form with neural attention}.
\newblock \bibinfo{journal}{Proceedings of the 54th Annual Meeting of the
  Association for Computational Linguistics (Volume 1: Long Papers)} \URLprefix
  \url{http://dx.doi.org/10.18653/v1/P16-1004},
  \DOIprefix\doi{10.18653/v1/p16-1004}.
\bibitem[{Elliott et~al.(2017)Elliott, Frank, Barrault, Bougares and
  Specia}]{Elliott_2017}
\bibinfo{author}{Elliott, D.}, \bibinfo{author}{Frank, S.},
  \bibinfo{author}{Barrault, L.}, \bibinfo{author}{Bougares, F.},
  \bibinfo{author}{Specia, L.}, \bibinfo{year}{2017}.
\newblock \bibinfo{title}{Findings of the second shared task on multimodal
  machine translation and multilingual image description}.
\newblock \bibinfo{journal}{Proceedings of the Second Conference on Machine
  Translation} \DOIprefix\doi{10.18653/v1/w17-4718}.
\bibitem[{Elliott et~al.(2016)Elliott, Frank, Sima’an and Specia}]{multi30k}
\bibinfo{author}{Elliott, D.}, \bibinfo{author}{Frank, S.},
  \bibinfo{author}{Sima’an, K.}, \bibinfo{author}{Specia, L.},
  \bibinfo{year}{2016}.
\newblock \bibinfo{title}{Multi30k: Multilingual english-german image
  descriptions}.
\newblock \bibinfo{journal}{Proceedings of the 5th Workshop on Vision and
  Language} \DOIprefix\doi{10.18653/v1/w16-3210}.
\bibitem[{Ge and Mooney(2005)}]{10.5555/1706543.1706546}
\bibinfo{author}{Ge, R.}, \bibinfo{author}{Mooney, R.J.}, \bibinfo{year}{2005}.
\newblock \bibinfo{title}{A statistical semantic parser that integrates syntax
  and semantics}, in: \bibinfo{booktitle}{Proceedings of the Ninth Conference
  on Computational Natural Language Learning}, \bibinfo{publisher}{Association
  for Computational Linguistics}, \bibinfo{address}{USA}. p.
  \bibinfo{pages}{9–16}.
\bibitem[{Haarnoja et~al.(2018)Haarnoja, Zhou, Abbeel and
  Levine}]{haarnoja2018soft}
\bibinfo{author}{Haarnoja, T.}, \bibinfo{author}{Zhou, A.},
  \bibinfo{author}{Abbeel, P.}, \bibinfo{author}{Levine, S.},
  \bibinfo{year}{2018}.
\newblock \bibinfo{title}{Soft actor-critic: Off-policy maximum entropy deep
  reinforcement learning with a stochastic actor}, in:
  \bibinfo{booktitle}{International Conference on Machine Learning},
  \bibinfo{organization}{PMLR}. pp. \bibinfo{pages}{1861--1870}.
\bibitem[{Helcl and Libovick{\'{y}}(2017)}]{NeuralMonkey:2017}
\bibinfo{author}{Helcl, J.}, \bibinfo{author}{Libovick{\'{y}}, J.},
  \bibinfo{year}{2017}.
\newblock \bibinfo{title}{Neural monkey: An open-source tool for sequence
  learning}.
\newblock \bibinfo{journal}{The Prague Bulletin of Mathematical Linguistics} ,
  \bibinfo{pages}{5--17}\DOIprefix\doi{10.1515/pralin-2017-0001}.
\bibitem[{Hochreiter and Schmidhuber(1997)}]{lstm}
\bibinfo{author}{Hochreiter, S.}, \bibinfo{author}{Schmidhuber, J.},
  \bibinfo{year}{1997}.
\newblock \bibinfo{title}{Long short-term memory}.
\newblock \bibinfo{journal}{Neural Comput.} \bibinfo{volume}{9},
  \bibinfo{pages}{1735–1780}.
\newblock \DOIprefix\doi{10.1162/neco.1997.9.8.1735}.
\bibitem[{Jabi et~al.(2019)Jabi, Pedersoli, Mitiche and Ayed}]{jabi2019deep}
\bibinfo{author}{Jabi, M.}, \bibinfo{author}{Pedersoli, M.},
  \bibinfo{author}{Mitiche, A.}, \bibinfo{author}{Ayed, I.B.},
  \bibinfo{year}{2019}.
\newblock \bibinfo{title}{Deep clustering: On the link between discriminative
  models and k-means}.
\newblock \bibinfo{journal}{IEEE Transactions on Pattern Analysis and Machine
  Intelligence} .
\bibitem[{Jia and Liang(2016)}]{Jia_2016}
\bibinfo{author}{Jia, R.}, \bibinfo{author}{Liang, P.}, \bibinfo{year}{2016}.
\newblock \bibinfo{title}{Data recombination for neural semantic parsing}.
\newblock \bibinfo{journal}{Proceedings of the 54th Annual Meeting of the
  Association for Computational Linguistics (Volume 1: Long Papers)}
  \DOIprefix\doi{10.18653/v1/p16-1002}.
\bibitem[{Karrupusamy(2020)}]{Karrupusamy2020EffectiveTS}
\bibinfo{author}{Karrupusamy, P.}, \bibinfo{year}{2020}.
\newblock \bibinfo{title}{Effective test suite optimization for improving the
  coverage standards using hybrid wrapper filter-memetic algorithm}.
\newblock \bibinfo{journal}{Journal of Social and Clinical Psychology}
  \bibinfo{volume}{2}, \bibinfo{pages}{83--91}.
\bibitem[{Kate and Mooney(2006)}]{kate-mooney-2006-using}
\bibinfo{author}{Kate, R.J.}, \bibinfo{author}{Mooney, R.J.},
  \bibinfo{year}{2006}.
\newblock \bibinfo{title}{Using string-kernels for learning semantic parsers},
  in: \bibinfo{booktitle}{Proceedings of the 21st International Conference on
  Computational Linguistics and 44th Annual Meeting of the Association for
  Computational Linguistics}, \bibinfo{publisher}{Association for Computational
  Linguistics}, \bibinfo{address}{Sydney, Australia}. pp.
  \bibinfo{pages}{913--920}.
\bibitem[{Liang et~al.(2011)Liang, Jordan and Klein}]{liang-etal-2011-learning}
\bibinfo{author}{Liang, P.}, \bibinfo{author}{Jordan, M.},
  \bibinfo{author}{Klein, D.}, \bibinfo{year}{2011}.
\newblock \bibinfo{title}{Learning dependency-based compositional semantics},
  in: \bibinfo{booktitle}{Proceedings of the 49th Annual Meeting of the
  Association for Computational Linguistics: Human Language Technologies},
  \bibinfo{publisher}{Association for Computational Linguistics},
  \bibinfo{address}{Portland, Oregon, USA}. pp. \bibinfo{pages}{590--599}.
\bibitem[{Luong et~al.(2015)Luong, Pham and Manning}]{luong2015effective}
\bibinfo{author}{Luong, M.T.}, \bibinfo{author}{Pham, H.},
  \bibinfo{author}{Manning, C.D.}, \bibinfo{year}{2015}.
\newblock \bibinfo{title}{Effective approaches to attention-based neural
  machine translation}.
\newblock \href{http://arxiv.org/abs/1508.04025}{{\tt arXiv:1508.04025}}.
\bibitem[{Papineni et~al.(2002)Papineni, Roukos, Ward and
  Zhu}]{papineni-etal-2002-bleu}
\bibinfo{author}{Papineni, K.}, \bibinfo{author}{Roukos, S.},
  \bibinfo{author}{Ward, T.}, \bibinfo{author}{Zhu, W.J.},
  \bibinfo{year}{2002}.
\newblock \bibinfo{title}{{B}leu: a method for automatic evaluation of machine
  translation}, in: \bibinfo{booktitle}{Proceedings of the 40th Annual Meeting
  of the Association for Computational Linguistics},
  \bibinfo{publisher}{Association for Computational Linguistics},
  \bibinfo{address}{Philadelphia, Pennsylvania, USA}. pp.
  \bibinfo{pages}{311--318}.
\newblock \DOIprefix\doi{10.3115/1073083.1073135}.
\bibitem[{Raffin et~al.(2019)Raffin, Hill, Ernestus, Gleave, Kanervisto and
  Dormann}]{stable-baselines3}
\bibinfo{author}{Raffin, A.}, \bibinfo{author}{Hill, A.},
  \bibinfo{author}{Ernestus, M.}, \bibinfo{author}{Gleave, A.},
  \bibinfo{author}{Kanervisto, A.}, \bibinfo{author}{Dormann, N.},
  \bibinfo{year}{2019}.
\newblock \bibinfo{title}{Stable baselines3}.
\newblock
  \bibinfo{howpublished}{\url{https://github.com/DLR-RM/stable-baselines3}}.
\bibitem[{Sutton and Barto(2018)}]{rlbook}
\bibinfo{author}{Sutton, R.S.}, \bibinfo{author}{Barto, A.G.},
  \bibinfo{year}{2018}.
\newblock \bibinfo{title}{Reinforcement Learning: An Introduction}.
\newblock \bibinfo{publisher}{A Bradford Book}, \bibinfo{address}{Cambridge,
  MA, USA}.
\bibitem[{Wong and Mooney(2006)}]{wong-mooney-2006-learning}
\bibinfo{author}{Wong, Y.W.}, \bibinfo{author}{Mooney, R.},
  \bibinfo{year}{2006}.
\newblock \bibinfo{title}{Learning for semantic parsing with statistical
  machine translation}, in: \bibinfo{booktitle}{Proceedings of the Human
  Language Technology Conference of the {NAACL}, Main Conference},
  \bibinfo{publisher}{Association for Computational Linguistics},
  \bibinfo{address}{New York City, USA}. pp. \bibinfo{pages}{439--446}.
\newblock \URLprefix \url{https://aclanthology.org/N06-1056}.
\bibitem[{Yang et~al.(2017)Yang, Fu, Sidiropoulos and Hong}]{yang2017towards}
\bibinfo{author}{Yang, B.}, \bibinfo{author}{Fu, X.},
  \bibinfo{author}{Sidiropoulos, N.D.}, \bibinfo{author}{Hong, M.},
  \bibinfo{year}{2017}.
\newblock \bibinfo{title}{Towards k-means-friendly spaces: Simultaneous deep
  learning and clustering}, in: \bibinfo{booktitle}{international conference on
  machine learning}, \bibinfo{organization}{PMLR}. pp.
  \bibinfo{pages}{3861--3870}.
\bibitem[{Yang and Whinston(2022)}]{yang2021identifying}
\bibinfo{author}{Yang, Y.}, \bibinfo{author}{Whinston, A.},
  \bibinfo{year}{2022}.
\newblock \bibinfo{title}{Identifying mislabeled images in supervised learning
  utilizing autoencoder}, in: \bibinfo{editor}{Arai, K.} (Ed.),
  \bibinfo{booktitle}{Proceedings of the Future Technologies Conference (FTC)
  2021, Volume 2}, \bibinfo{publisher}{Springer International Publishing},
  \bibinfo{address}{Cham}. pp. \bibinfo{pages}{266--282}.
\bibitem[{Yang and Xue(2023)}]{yang2021training}
\bibinfo{author}{Yang, Y.}, \bibinfo{author}{Xue, Z.}, \bibinfo{year}{2023}.
\newblock \bibinfo{title}{Training heterogeneous features in sequence
  to sequence tasks: Latent enhanced multi-filter seq2seq model}, in:
  \bibinfo{editor}{Arai, K.} (Ed.), \bibinfo{booktitle}{Intelligent Systems and
  Applications}, \bibinfo{publisher}{Springer International Publishing},
  \bibinfo{address}{Cham}. pp. \bibinfo{pages}{103--117}.
\bibitem[{Zelle and Mooney(1996)}]{zelle_mooney}
\bibinfo{author}{Zelle, J.M.}, \bibinfo{author}{Mooney, R.J.},
  \bibinfo{year}{1996}.
\newblock \bibinfo{title}{Learning to parse database queries using inductive
  logic programming}, in: \bibinfo{booktitle}{Proceedings of the Thirteenth
  National Conference on Artificial Intelligence - Volume 2},
  \bibinfo{publisher}{AAAI Press}. p. \bibinfo{pages}{1050–1055}.

\end{thebibliography}
\bibliographystyle{elsarticle-harv}

\end{document}